\documentclass{IEEEtran}

\usepackage{graphicx}
\usepackage{lineno}
\modulolinenumbers[5]
\usepackage[cmex10]{amsmath}
\usepackage{amsfonts}

\usepackage{color}
\usepackage{algorithm}
\usepackage{algorithmic}
\usepackage{array}
\usepackage{multirow}
\usepackage{booktabs}
\usepackage{subfigure}
\usepackage{bm}

\allowdisplaybreaks[4]
\usepackage{underscore}
\usepackage{stfloats}
\usepackage{float}
\usepackage{lipsum}
\usepackage{cite}
\usepackage{amssymb}
\usepackage{amsmath}
\usepackage{mathrsfs}
\usepackage[margin=0.75in]{geometry}
\usepackage{amsthm}

\newtheorem{definition}{Definition}

\makeatletter
\begin{document}

\title{
  World Model-Based Learning for Long-Term Age of Information Minimization in Vehicular Networks


\author{
\IEEEauthorblockN{
Lingyi Wang\IEEEauthorrefmark{1}, 
Rashed Shelim\IEEEauthorrefmark{1},  
Walid Saad\IEEEauthorrefmark{1},
and Naren Ramakrishnan\IEEEauthorrefmark{2}\\
\IEEEauthorblockA{\IEEEauthorrefmark{1}Bradley Department of Electrical and Computer Engineering, Virginia Tech, Alexandria, VA, 22305, USA}\\
\IEEEauthorblockA{\IEEEauthorrefmark{2}Department of Computer Science, Virginia Tech, Alexandria, VA, 22305, USA}\\
\IEEEauthorblockA{Emails: \{lingyiwang, rasheds, walids, naren\}@vt.edu
}}
}
}
\thispagestyle{empty}
\maketitle
\thispagestyle{empty}

\begin{abstract}
  Traditional reinforcement learning (RL)-based learning approaches for wireless networks rely on expensive trial-and-error mechanisms and real-time feedback based on extensive environment interactions, which leads to low data efficiency and short-sighted policies. These limitations become particularly problematic in complex, dynamic networks with high uncertainty and long-term planning requirements. To address these limitations, in this paper, a novel world model-based learning framework is proposed to minimize packet-completeness-aware age of information (CAoI) in a vehicular network. Particularly, a challenging representative scenario is considered pertaining to a millimeter-wave (mmWave) vehicle-to-everything (V2X) communication network, which is characterized by high mobility, frequent signal blockages, and extremely short coherence time. Then, a world model framework is proposed to jointly learn a dynamic model of the mmWave V2X environment and use it to imagine trajectories for learning how to perform link scheduling. In particular, the long-term policy is learned in differentiable imagined trajectories instead of environment interactions. Moreover, owing to its imagination abilities, the world model can jointly predict time-varying wireless data and optimize link scheduling in real-world wireless and V2X networks. Thus, during intervals without actual observations, the world model remains capable of making efficient decisions. 
  Extensive experiments are performed on a realistic simulator based on Sionna that integrates physics-based end-to-end channel modeling, ray-tracing, and scene geometries with material properties.
  Simulation results show that the proposed world model achieves a significant improvement in data efficiency, and achieves 26\% improvement and 16\% improvement in CAoI respectively compared to the model-based RL (MBRL) method and the model-free RL (MFRL) method.
\end{abstract}

\IEEEpeerreviewmaketitle

\section{Introduction}
Over the last decade, advances in reinforcement learning (RL) techniques, both model-free RL (MFRL) and model-based RL (MBRL), have led to the proliferation of learning-based optimization across a broad range of wireless networking use cases. These approaches provide intelligent and efficient solutions for tasks such as real-time beamforming \cite{9612729}, dynamic link access \cite{9930939}, and resource scheduling \cite{wang2023adaptive}. Compared to traditional optimization approaches, learning-based approaches learn directly from the data, adapt to environment variations, and avoid the need for handcrafted rules or static models. However, despite this promising potential, the application of RL to network optimization still faces several significant limitations. Particularly, complex and dynamic communication environments, such as those with rapid topology changes, uncertain channels, and heterogeneous devices, pose significant challenges for the RL process. Those challenges include environment uncertainty, low sample efficiency, limited real-time data availability, and the need for long-term planning. For instance, the works in \cite{9612729,9930939,wang2023adaptive} investigated MFRL-based approaches for wireless resource allocation, where the policy was learned from instant feedback obtained by real environment interactions. However, these approaches heavily rely on trial-and-error mechanisms with short-sighted reward signals, thus they are data inefficient. In \cite{liao2019model,9852968,park2023model}, the authors focused on MBRL-based approaches that improved data efficiency by learning the wireless dynamics. However, MBRL techniques typically rely on high-dimensional original observation spaces and non-differentiable policy learning through sampling-based trajectory evaluation. Hence, MBRL approaches are unable to address the credit assignment problem that attributes delayed rewards back to the earlier actions. 

To address the aforementioned limitations of RL approaches, recent works in the machine learning community proposed \emph{world model}-based learning frameworks \cite{10929033,hafner2019dream,hafner2023mastering,wang2025dmwm}, which decouple environment modeling from policy learning. By learning a predictive model of the environment dynamics and uncertainty, a world model enables agents to imagine the future impact of current actions in a compact latent space. Those imagined trajectories are differentiable, and, thus, a long-horizon policy can be learned by accurately attributing future rewards to earlier decisions without interactions and feedback from the actual environment. World models have been widely used in learning policy from visual data and have shown significant improvement in task performance.
However, the existing world models \cite{hafner2019dream,hafner2023mastering,wang2025dmwm} can not be directly used in wireless networks \cite{10929033}. The observations of the wireless environment, such as channel state information, antenna angles and path delay, are high-dimensional, sparse, and heavily noisy physical quantities. It is challenging for existing world model approaches to explore complex spatio-temporal structures from wireless data. Moreover, wireless data is highly dynamic within extremely short time slots caused by multipath propagation, blockages, and mobility. These features require the world model to provide highly precise predictions at finer temporal scales.

The main contribution of this paper is a novel world model framework that can effectively model the uncertainty and dynamics of wireless networks, enhance data efficiency, and endow the network with long-term planning ability. We consider a particularly challenging scenario pertaining to a millimeter-wave (mmWave) vehicle-to-everything (V2X) communication network \cite{tunc2021mitigating}, that exhibits high mobility, frequent link blockages, and temporal dependency of states. These features of a V2X network require real-time, reliable, and forward-looking optimization. To address these challenges, we propose a world model-based learning framework to minimize the packet-completeness-aware age of information (CAoI) over a long horizon by optimizing link pairs. The proposed world model jointly learns the dynamic model of the mmWave V2X network and uses it for long-term trajectory prediction and link scheduling. By integrating a recurrent state-space model (RSSM) with an actor-critic policy module, the proposed framework enables sample-efficient and long-term planning ability for wireless networks, even with high dynamics, uncertainty and sparse rewards. In particular, long-term link scheduling is learned in differentiable imagination trajectories instead of environment interactions. Moreover, the imagination ability of the world model is further utilized to jointly predict time-varying wireless data and optimize link scheduling in real-world mmWave V2X networks without real-time collection of wireless data. We conduct experiments on a realistic simulator based on Sionna that integrates physics-based end-to-end channel modeling, ray-tracing, and detailed scene geometries with material properties. Simulation results show that the proposed world model achieves a significant improvement in data efficiency, and achieves 26\% improvement and 16\% improvement in CAoI respectively compared to the MBRL method and the MFRL method.


\section{mmWave V2X Communication System}
We consider a mmWave V2X network composed of one roadside unit (RSU) $u$ and a set $\mathcal{V}$ of $V$ mobile vehicles, as shown in Fig. 1. The network comprises both vehicle-to-infrastructure (V2I) and vehicle-to-vehicle (V2V). Let $\mathcal{M}[t]$ and $\mathcal{Z}[t]$ respectively be the time-varying sets of V2I and V2V link pairs. The V2V/V2I links share a total bandwidth $B$. We use narrow and directional beams, and, thus, there is no interference in the V2X network \cite{tunc2021mitigating}.
We consider a time-slotted system, where each timeslot is indexed by \( t \) and has a fixed duration $\xi$. Each vehicle operates in a half-duplex communication mode, where it can establish only one communication link during a timeslot $t$, and is unable to transmit and receive data simultaneously.

\begin{figure}
  \centering
  \includegraphics[scale=0.62]{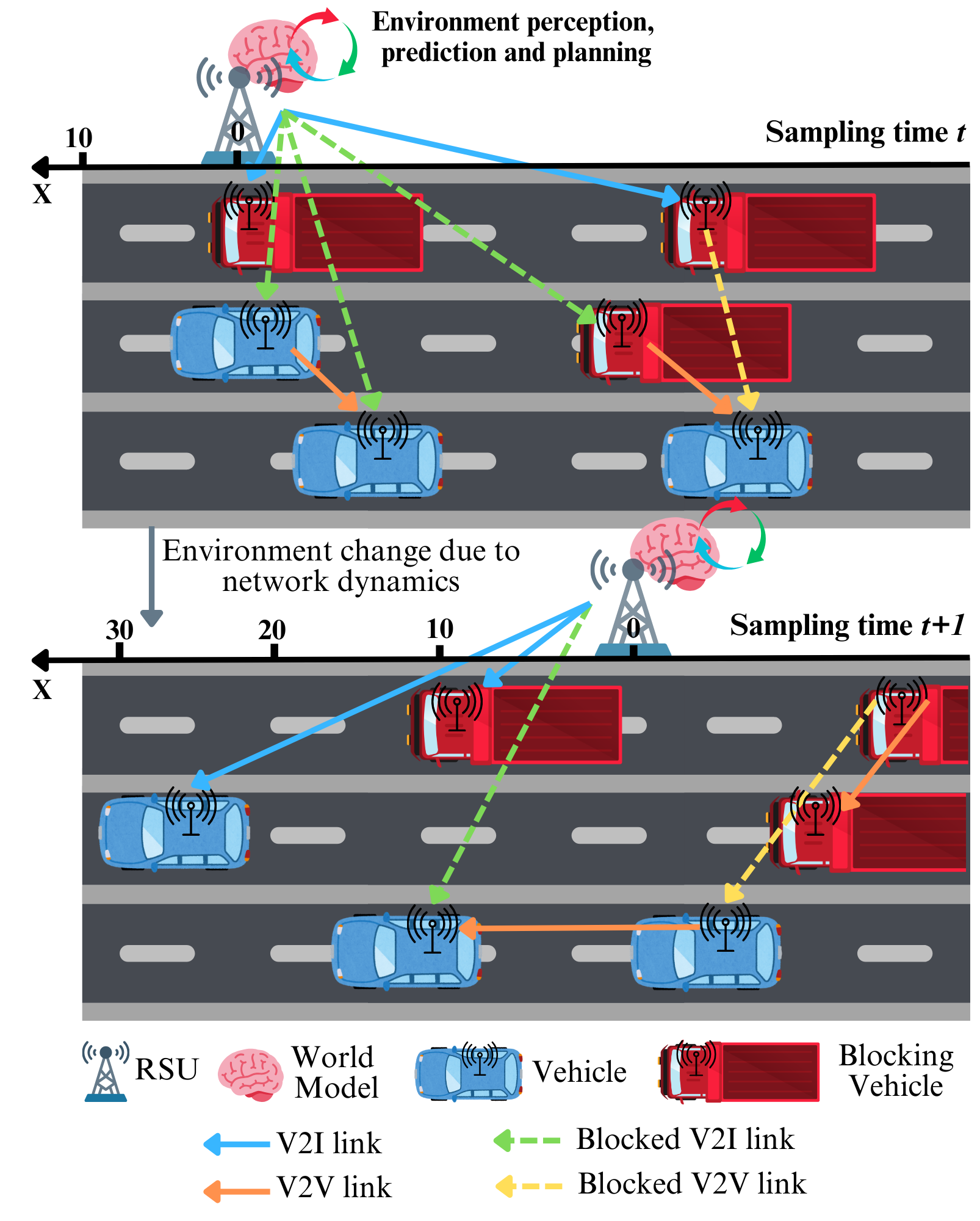}
  \vspace{-0.2cm}
  \caption{The mmWave V2X communication system with the world model.}
  \vspace{-0.3cm}
\end{figure}

\vspace{-0.2cm}
\subsection{Transmission Model}
Let $g_{ij}[t]$ be the mmWave V2X channel gain of timeslot $t$ from the transmitter of link $i$ to the receiver of link $j$. $g_{ij}[t]$ is characterized by high path loss, multipath propagation, dynamic blockages, and Doppler shifts.
The data rate, in packets per timeslot $t$, for V2I pair $m \in \mathcal{M}[t]$ and V2V pair $z \in \mathcal{Z}[t]$, respectively, is given by
\vspace{-0.1cm}
\begin{equation}
  \begin{aligned}
  &R^{\mathrm{V2V}}_{z}[t]=\frac{B \xi}{S}  \log\left(1+\frac{P_\mathcal{V}g_{zz}[t]}{N_0 B}\right), \\
  &R^{\mathrm{V2I}}_{m}[t]=\frac{B \xi}{S}  \log\left(1+
  \frac{P_u g_{mm}[t]}{N_0 B}\right),
  \end{aligned}
\end{equation}
where $S$ is the size of each packet, $N_0$ is the power spectral density of additive white Gaussian noise, and $P_u$ and $P_\mathcal{V}$ are respectively the transmit power of the RSU and each vehicle.

In mmWave V2X communication, blockages caused by high-speed mobile vehicles, buildings, and other obstacles significantly impact signal propagation and can lead to disruptions in both V2V and V2I links. To model the blockage effect, we consider the Fresnel zone obstruction \cite{9838711}, path loss variations, and environmental dynamic characteristics in our blockage model. 
The first Fresnel zone radius determines the critical region for obstruction as
\begin{equation}
r_\mathrm{F}= \sqrt{\frac{\lambda d_{kb} d_{br}}{d_{kr}}},
\end{equation}
where $d_{kb}$ and $d_{br}$ are respectively the distances from the blocking vehicle to the transmitter and receiver, with $d_{kr} = d_{kb} + d_{br}$ being the total link distance. A blockage occurs when the height of the blocking vehicle $h_b$ exceeds the effective Fresnel height $h_F$, which is given by
\begin{equation}
h_\mathrm{F}= h[t] + \frac{(h_r - h[t]) d_{kb}}{d_{kr}} - 0.6 r_F,
\end{equation}
where $h[t]$ and $h_r$ respectively represent the antennas heights of the transmitter and receiver.
Assume the vehicle heights follow a Gaussian distribution $h_b \sim \mathcal{N}(\mu_b, \sigma_b^2)$, the probability of blockage will be
\begin{equation}
P^{\mathrm{Block}}_\mathrm{F}= Q\left( \frac{h_\mathrm{F}- \mu_b}{\sigma_b} \right),
\end{equation}
where $Q(x)$ is the Gaussian Q-function. For multiple blocking vehicles, the number of vehicles follows a Poisson point process with vehicle density $\lambda_v$, and the LoS probability of V2V and V2I links are respectively given by
\vspace{-0.1cm}
\begin{equation}
  \begin{aligned}
  &P_{\text{LoS}}^{{\text{V2V}}}(d_{kr}) = e^{-\lambda_v d_{kr} P^{\mathrm{Block}}_\mathrm{F}},\\
  &P_{\text{LoS}}^{{\text{V2I}}}(d_{kr}) = P_{\text{LoS}}^{\text{3GPP}}(d_{kr}) \cdot e^{-\lambda_v d_{kr} P^{\mathrm{Block}}_\mathrm{F}},
  \end{aligned}
\end{equation}
where $P_{\text{LoS}}^{\text{3GPP}}(d_{kr}) = e^{-\beta d_{kr}}$ is the 3GPP empirical model that captures urban blockages from buildings, and $\beta$ is a factor that depends on the physical environment.

\begin{figure*}
  \begin{equation}\label{eq2}
    \begin{aligned}
    &\tilde{G}_{v,m}[t] = \mathbb{I}(\lfloor R^{\mathrm{V2I}}_{m}[t] \rfloor \geq C_u) G_u[t] + \mathbb{I}(\lfloor R^{\mathrm{V2I}}_{m}[t]\rfloor < C_u)\left[\frac{\lfloor R^{\mathrm{V2I}}_{m}[t] \rfloor}{C_u}G_u[t] + \frac{C_u - \lfloor R^{\mathrm{V2I}}_{m}[t] \rfloor}{C_u}A_v[t]\right],\\
    &\tilde{G}_{v,z}[t] = \mathbb{I}(\lfloor R^{\mathrm{V2V}}_{z}[t] \rfloor \geq C_{v^{\prime}}[t]) \left[\frac{C_{v^{\prime}}[t]}{C_u}G_{v^{\prime}}[t] + \frac{C_u - C_{v^{\prime}}[t]}{C_u}A_v[t]\right] \\
    & \quad \quad \quad \quad \quad \quad \quad \quad \quad \quad \quad \quad + \mathbb{I}(\lfloor R^{\mathrm{V2V}}_{z}[t] \rfloor < C_{v^{\prime}}[t])\left[\frac{\lfloor R^{\mathrm{V2V}}_{z}[t] \rfloor}{C_u}G_{v^{\prime}}[t] + \frac{C_u - \lfloor R^{\mathrm{V2I}}_{m}[t] \rfloor}{C_u}A_v[t]\right].\\
  \end{aligned}
  \end{equation}
  \hrulefill
  \vspace{-0.2cm}
\end{figure*}

\vspace{-0.4cm}
\subsection{CAoI Metric}
The RSU must transmit road information that consists of $C_u$ packets during each timeslot. This information includes real-time data such as traffic signal timing, roadside sensor messages and emergency warning. We consider a practical broadcast scenario, in which the RSU must enable vehicles to receive complete, up-to-date data for both driving efficiency and safety. The age of information (AoI) is usually used to quantify end-to-end latency but ignore reliability of highly-dynamic mmWave V2X links. When blockages or severe path loss occur in mmWave networks, packets can be truncated and partially received. Hence, in the next definition, we introduce the concept of a CAoI that scales the AoI by each link's transmission rate to more accurately capture the delivered information freshness. We define CAoI as follows. 

\begin{definition}
The \emph{CAoI} $A_v[t+1]$ of a vehicle $v\in \mathcal{V}$ at the timeslot $t+1$ in the V2X communication network is given by
\begin{equation}
  \begin{aligned}
    A_v[t+1] = 
\begin{cases} 
  t -  \tilde{G}_{v,m}[t] + 1, &v \text{ receives from V2I pair } m,  \\ 
  t - \tilde{G}_{v,z}[t] + 1, & v \text{ receives from V2V pair } z,  \\ 
A_v[t] + 1, & \text{otherwise}.
\end{cases}
  \end{aligned}
\end{equation} 
\end{definition}

\noindent In (7), $\tilde{G}_{v,m}[t]$ and $\tilde{G}_{v,z}[t]$ are given by (\ref{eq2}), and they respectively represent the CAoI update of vehicle $v$ over V2I link $m$ and V2V link $z$. In (\ref{eq2}), $C_{v^{\prime}}$ is the number of expected packets from the vehicle transmitter $v^{\prime}$, $G_{v,m}[t]$ is the updated information age by packets delivered from the V2V link $m$, $G_{v,z}[t]$ is the updated information age by packets delivered from the V2V link $z$, and
$G_{u}[t]$ and $G_{v^{\prime}}[t]$ are, respectively, the CAoI of the RSU and vehicle transmitter $v^{\prime}$.
The indicator function \(\mathbb{I}(x)\) is a binary-valued function that equals to 1 if the condition \( x \) holds true and 0 otherwise.
\vspace{0.2cm}

\vspace{-0.3cm}
\subsection{CAoI Minimization Problem}
The goal of the network is to minimize its average CAoI by providing reliable link scheduling during a time period $T$ for packet transmission, as follows:
\begin{subequations}\label{opt}
  \begin{align}
    &\min_{\left\{\mathcal{M}[t], \mathcal{Z}[t]\right\}} \frac{1}{T} \sum_{t=1}^{T} \sum_{v}^{V} A_v[t] \\ 
   \text{ s.t. }  
   & A_v[t] \leq A^{\mathrm{max}}, \forall v \in \mathcal{V}, \\
   & \Phi_{\cap}(\mathcal{M}[t],\mathcal{Z}[t]) = \emptyset, 
  \end{align}
\end{subequations}
where $A^{\mathrm{max}}$ is the maximum age tolerance, $\Phi_{\cap}(\mathcal{M}[t],\mathcal{Z}[t])$ represents the shared link node (transmitter or receiver) set between the V2I link set $\mathcal{M}[t]$ and the V2V link set $\mathcal{Z}[t]$.
It is challenging to optimize the link scheduling in (8) due to the high mobility of the V2X network, mmWave blockages, and the dynamic temporal impact of link scheduling. Particularly, the current link scheduling influences the future system states by updating CAoI. Hence, the scheduling optimization needs to jointly consider the CAoI states of vehicles and the reliability of links affected by time-varying blockages and path loss during multiple time steps in the future. In this context, greedy or short-term optimal solutions provided by traditional RL schemes can be ineffective since they can not attribute delayed rewards back to the earlier actions. Moreover, since it is costly to obtain the real-world wireless data, traditional RL schemes also exhibit low data efficiency. Hence, a more efficient learning framework is required.

\section{World Model For Long-Term Prediction and Link Scheduling}
In this section, we introduce a novel world model framework, shown in Fig. 2, to solve the CAoI minimization (\ref{opt}). A world model framework is proposed to solve (8) due to its data efficiency and long-term planning ability. 

\begin{figure*}\label{f2}
  \centering
  \includegraphics[scale=0.85]{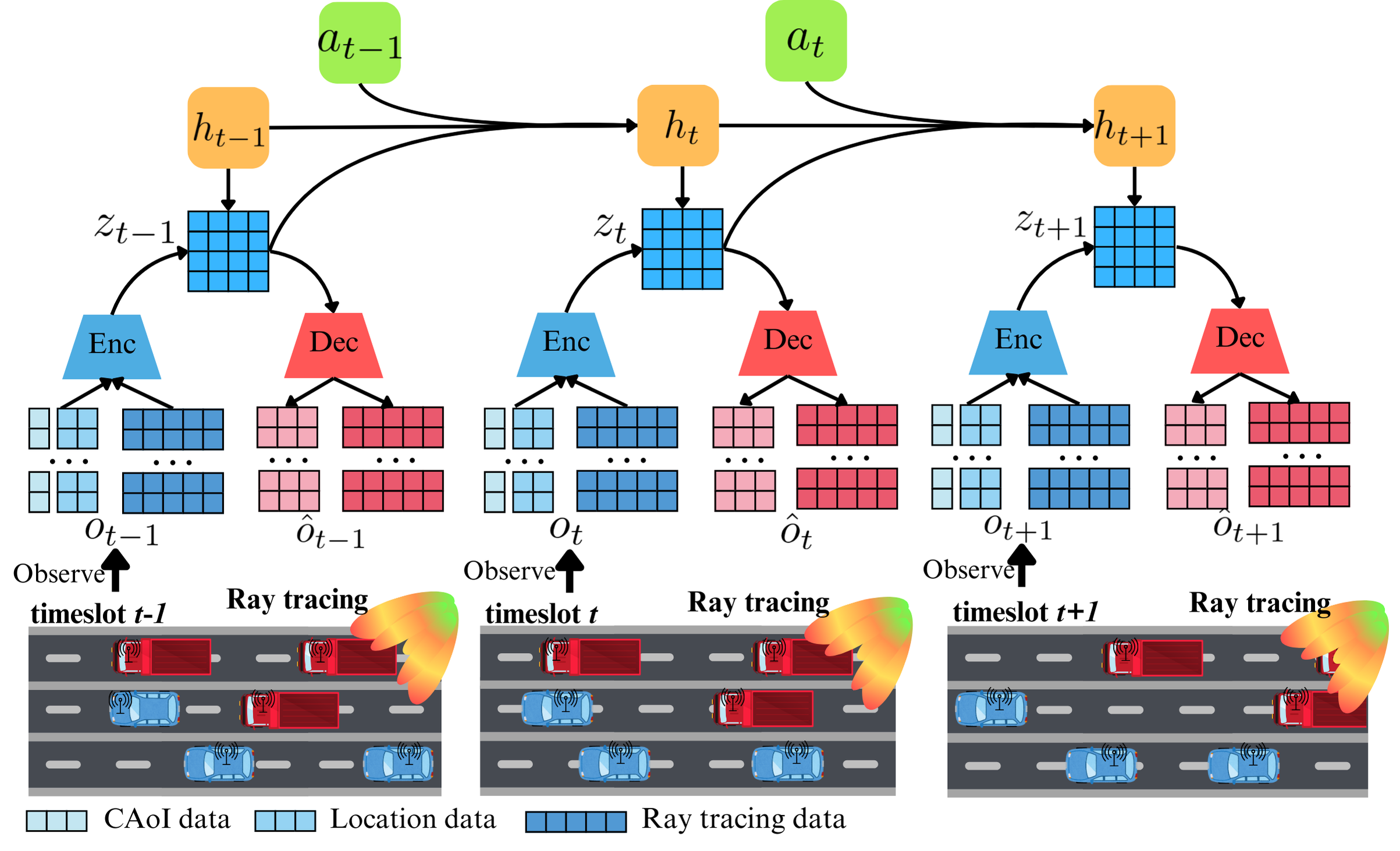}
  \vspace{-0.4cm}
  \caption{Learning a world model for V2X communication networks based on the location data and the ray tracing data.}
  \vspace{-0.5cm}
\end{figure*}

\vspace{-0.5cm}
\subsection{Framework for Learning Dynamics}
\vspace{-0.05cm}
In our world model, wireless factors that include CAoI $\{A_v[t]\}$, ray-tracing data $R[t]$ and vehicle locations $L[t]$ serve as observations $o[t] = \{\{A_v[t]\}_{v=1}^{V},R[t],L[t]\}$. Particularly, CAoI enables the agent to capture the information freshness of each vehicle, and vehicle locations provide the spatial context of geometric relationships among transmitters and receivers. Ray-tracing data captures the instantaneous multipath impact, blockage events, and Doppler effects that drive rapid channel changes in mmWave bands. The action of the world model is to do link scheduling, given by $a[t]=\left\{\mathcal{M}[t], \mathcal{Z}[t]\right\}$.

We adopt RSSM \cite{hafner2019dream,hafner2023mastering,wang2025dmwm} as a backbone for our world model. In particular, RSSM combines latent space modeling and recurrent structures along with variational inference to learn the dynamics of the mmWave V2X network. For imagined trajectories of wireless data over a long horizon, traditional recurrent neural network (RNN)-driven schemes face the state drift issue due to the recursive state update directly based on historical data, while RSSM can effectively alleviate this issue and improve the stability by explicitly modeling the environment uncertainty. The RSSM is represented by
\vspace{-0.1cm}
\begin{align*}
    \begin{array}{ll}
    \text{Deterministic state:}  &h[t] = f_\varphi\left(h[t-1], z[t-1], a[t-1]\right),\\
    \text{Encoder:}  &z[t] \sim q_\varphi\left(z[t] \mid h[t], o[t]\right), \\
    \text{Stochastic state:}  &\tilde{z}[t] \sim p_\varphi\left(\tilde{z}[t] \mid h[t]\right), \\
    \text{Reward predictor:}  &\tilde{r}[t] \sim p_\varphi\left(\tilde{r}[t] \mid h[t], z[t]\right), \\
    \text{Decoder:}  &\hat{o}[t] \sim p_\varphi\left(\hat{o}[t] \mid h[t], z[t]\right),
    \end{array}  
\end{align*} 
where $h[t]$ is the deterministic state, $z[t]$ is the stochastic state, $\hat{o}[t]$ is the recovered observations, $\tilde{z}[t]$ is the prediction of the stochastic state, and $\tilde{r}[t]$ is the prediction of the real-world reward during timeslot $t$. The loss function of the RSSM is expressed as $\mathcal{L}(\varphi) = \mathcal{L}_{\text {pred }}(\varphi) + \varpi_{1} \mathcal{L}_{\text {dyn }}(\varphi) + \varpi_{2}  \mathcal{L}_{\text {rep }}(\varphi)$ with the weight factors $\varpi_{1}$ and $\varpi_{2}$. The loss term $\mathcal{L}_{\text {pred }}(\varphi) = -\ln p_\varphi\left(\hat{o}[t] \mid z[t], h[t]\right)-\ln p_\varphi\left(\tilde{r}[t] \mid z[t], h[t]\right)$ ensures $z[t]$ caputures the features from wireless data $o[t]$ and the credit assignment $\tilde{r}[t]$. 
The loss terms $\mathcal{L}_{\text {dyn }}(\varphi) = \operatorname{KL}\left[\operatorname{sg}\left(q_\varphi\left(z[t] \mid h[t], o[t]\right)\right) \| p_\varphi\left(\tilde{z}[t] \mid h[t]\right)\right]$ and $\mathcal{L}_{\text {rep }}(\varphi) = \operatorname{KL}\left[q_\varphi\left(z[t] \mid h[t], o[t]\right) \| \operatorname{sg}\left(p_\varphi\left(\tilde{z}[t] \mid h[t]\right)\right)\right]$ ensure $z[t]$ and $h[t]$ to learn the network dynamics in a latent space, where $\operatorname{sg}(\cdot)$ represents the stop-gradient operator, and $\operatorname{KL}(\cdot)$ represents Kullback-Leibler divergence. 

\begin{figure}[t!]
  \centering
  \includegraphics[scale=0.82]{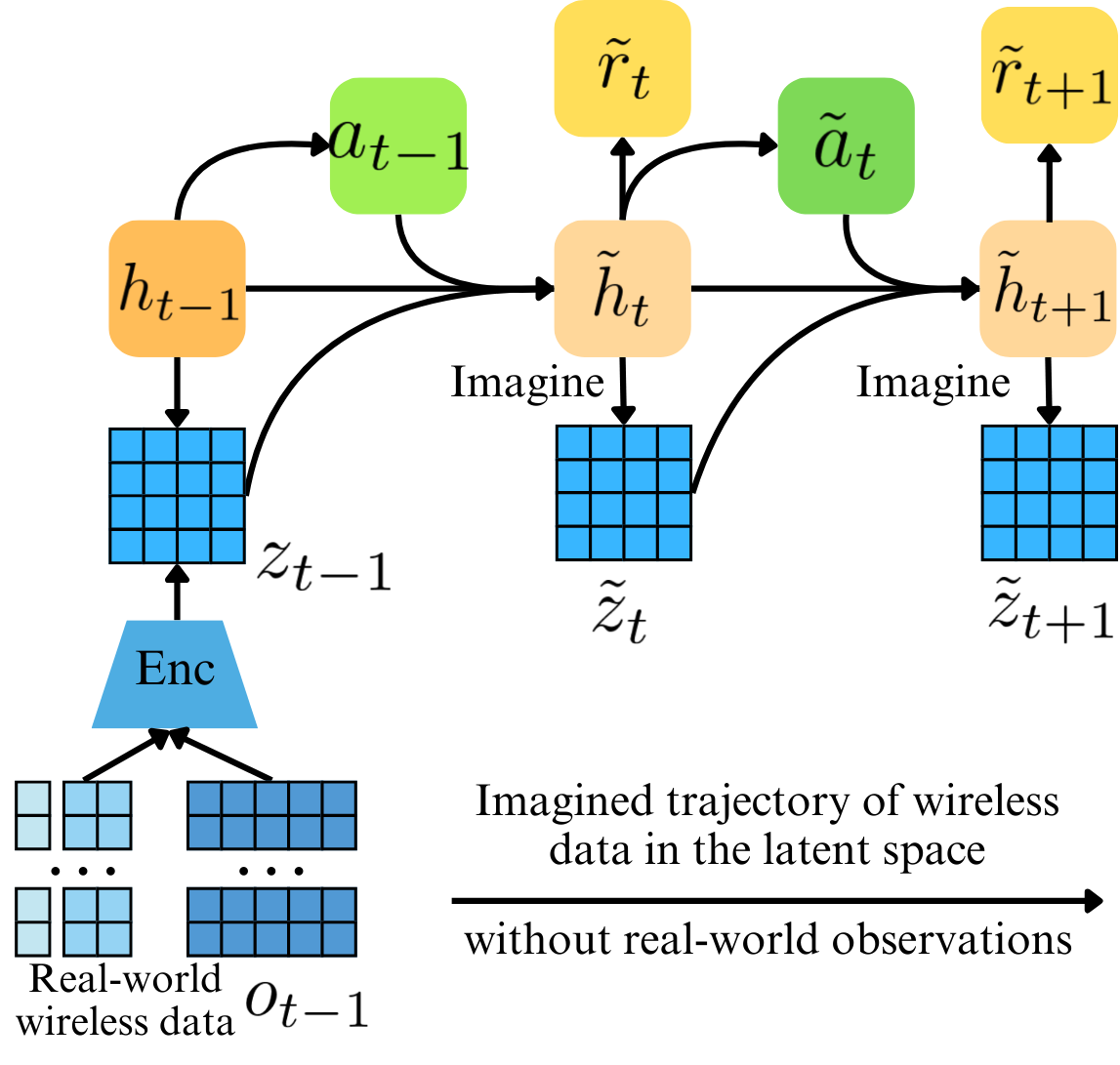}
    \vspace{-0.2cm}
  \caption{The imagination ability of the world model for both the actor-critic learning and the online joint prediction and scheduling.}
  \vspace{-0.5cm}
\end{figure}

\subsection{Imagination Ability for Learning Link Scheduling}
As shown in Fig. 3, the imagination ability of the world model is used to simulate the future trajectories of wireless data $\{\tilde{z}[t]\}$ in a latent space. It aims to learn long-term strategies in a data-efficient way without relying on high-cost interactions and expensive real-time feedback from the real-world V2X network. In particular, the prediction of the stochastic state is recurrently obtained by $\tilde{z}[t] \sim p_\varphi\left(\tilde{z}[t] \mid h[t]\right)$ and $h[t] = f_\varphi\left(h[t-1], \tilde{z}[t-1], \tilde{a}[t-1]\right)$. Then, an imagined trajectory of wireless network is formulated as $\tilde{\mathcal{J}}[t-1]=\left\{\tilde{s}[t:t+H],\tilde{a}[t:t+H],\tilde{r}[t:t+H]\right\}$, where state $\tilde{s}[t]=\left\{\tilde{z}[t],h[t]\right\}$ encodes wireless network data in the latent space, and $H$ is the horizon size of imagination.

We apply an actor-critic model to learn link scheduling in imagination. The actor-critic model involves two components: the actor component for policy learning and the critic component for state value estimation, represented by
\begin{equation}
  \begin{aligned}
  &\text{Actor model: } \tilde{a}[\tau] \sim q_{\theta}(\tilde{a}[\tau] \mid \tilde{s}[\tau]),\\ 
  &\text{Critic model: } v_\psi(\tilde{s}[\tau]) \approx \mathbb{E}_{q(\cdot \mid \tilde{s}[\tau])} \left( \sum_{t=\tau}^H \gamma^{t-\tau} \tilde{r}[\tau] \right),
  \end{aligned}
\end{equation}
where $\theta$ is the parameters of the actor, and $\psi$ is the parameters of the critic.
In the imagined trajectory $\tilde{\mathcal{J}}$ of the varying wireless network, the actor learns to maximize the return value with each link schedule, and the target of the critic model learns to evaluate the value with CAoI. The goal of the actor and the critic are respectively given by:
\begin{equation}
  \begin{aligned}
    &\theta^* = \max_\theta \mathbb{E}_{q_\phi, q_\theta} \left[ \sum_{\tau=t}^{t+H} V_\lambda(\tilde{s}[\tau]) \right] ,\\
    &\psi^* = \min_\psi \mathbb{E}_{q_\phi, q_\theta} \left[ \sum_{\tau=t}^{t+H} \frac{1}{2} \left( v_\psi(\tilde{s}[\tau]) - V_\lambda(\tilde{s}[\tau]) \right)^2 \right].
  \end{aligned}
  \end{equation}
The value $V_\lambda(\tilde{s}[\tau])$ with discount weight $\lambda$ is given by 
\vspace{-0.2cm}
\begin{equation}
  \begin{aligned}
    &V_\lambda(\tilde{s}[\tau]) = (1 - \lambda) \left( \sum_{n=1}^{H-1} \lambda^{n-1} V_n^N(\tilde{s}[\tau]) \right) + \lambda^{H-1} V_H^N(\tilde{s}[\tau]), \\
    &V_k^N(\tilde{s}[\tau]) = \mathbb{E}_{q_\phi, q_\theta} \left[ \sum_{n=\tau}^{h-1} \gamma^{n-\tau} \tilde{r}_n + \gamma^{h-\tau} v_\psi(\tilde{s}[h]) \right],
  \end{aligned}
\end{equation}
where $h = \min(\tau + k, t + H)$. Moreover, the actual reward $r[t]$ of the network during the timeslot $t$ is designed as
\begin{equation*}
  r[t] = - \frac{1}{V} \sum_{v}\left[ A_{v}[t] - \mathbb{I}(A_v[t] > A^{\mathrm{max}}) (A^{\mathrm{max}}-A_v[t])\right].
\end{equation*}

In real-world mmWave V2X networks, it is difficult and inefficient to obtain the real-time wireless data $\{o[t]\}$ within each small timeslot $t$ when the size of wireless data $\{o[t]\}$ is large. 
In this context, the imagination ability of the world model can be utilized for joint prediction of wireless data and link scheduling without the real-time data collection in the practical application, as illustraed in Fig. 3. Given the deterministic trajectory $\mathcal{J}[c]=\left\{s[1:c],a[1:c],r[1:c]\right\}$ that is collected from actual data and fed back over $c$ timeslots, the world model can generate the imagined trajectory $\tilde{\mathcal{J}}[c]=\left\{\tilde{s}[c+1:T],\tilde{a}[c+1:T],\tilde{r}[c+1:T]\right\}$ for the future few timeslots. Without the real observations $\{o[c+1:T]\}$ of the wireless data, the world model can still decide stable actions $a[c+1:T]$ based on the imagined network trajectory $\tilde{\mathcal{J}}[c]$.

In conclusion, the proposed world model-based learning approach solves (8) by first learning a dynamic model of the V2X environment with wireless data, and then using this model to imagine future trajectories. The world model addresses the following challenges: (a) It attributes delayed rewards back to earlier scheduling since the imagination is differentiable, (b) It is data efficient without real-world interaction for policy training, and (c) It learns to jointly optimize CAoI and link reliability over a long horizon $H$.

\begin{table}
  \centering
  \caption{Hyperparameter Setting}
  \begin{tabular}{|l|c|c|}
  \hline
  \textbf{Parameter} & \textbf{Symbol} & \textbf{Value} \\ \hline
  \multicolumn{3}{|l|}{\textbf{Environment}} \\ \hline
  Number of vehicles & $V$ & 8 \\ 
  Bandwidth & $B$ & 100 $~\mathrm{MHz}$ \\ 
  Packet size & $S$ & 5 $~\mathrm{MB}$ \\
  Frequency & $f_c$ & 26 $~\mathrm{GHz}$ \\ 
  Transmit power & $P_\mathcal{V}$, $P_u$ & 23 $~\mathrm{dBm}$ \\
  Timeslot duration & $\xi$ & 100 $~\mathrm{ms}$ \\
  Period & T & 100 \\ 
  Age tolerance & $A^{\mathrm{max}}$ & 8 \\
  Vehicle speed & --- & 15-20 $~\mathrm{m/s}$\\ \hline
  \multicolumn{3}{|l|}{\textbf{World model framework}} \\ \hline
  Seed episode &  --- & 5 \\ 
  Sequence length &  --- & 64 \\ 
  Training episodes &  --- & 1e3 \\
  Max episode length & --- & 100 \\
  Collect interval &  --- & 100 \\
  Replay buffer size & --- & 1e6 \\ 
  Batch size &  --- & 50 \\ 
  Imagination horizon & $H$ & 30 \\ 
  Stochastic state size & $|z[t]|$, $|\tilde{z}[t]|$ & 256 \\ 
  Deterministic state size & $|h[t]|$, $|h[t]|$ & 256 \\ 
  Activation layer function & --- & Relu \\ 
  Loss weights & $\varpi_{_{\text {dyn}}}$, $\varpi_{_{\text {rep}}}$ & 1 \\ 
  World model optimizer & --- & Adam ($\epsilon$ = 1e-4) \\ 

  Learning rate & $\eta_\psi$ & 1e-3 \\
  \hline
  \multicolumn{3}{|l|}{\textbf{Actor-critic for policy learning}} \\ \hline
  Exploration noise & --- & 0.3 \\
  Return lambda & $\lambda$ & 0.95 \\ 
  Planning horizon discount & $\gamma$ & 0.99 \\ 
  Actor-critic optimizer & --- & Adam ($\epsilon$ = 1e-4) \\ 
  Learning rate & $\eta_{\vartheta}$, $\eta_{\psi}$ & 1e-4 \\ 
  \hline
  \end{tabular} 
\end{table}

\vspace{-0.3cm}
\section{Simulation Results and Analysis}
For our simulations, we use Sionna \cite{sionna} to generate a realistic mmWave V2X scenario at operating frequency $28 \mathrm{~GHz}$, where physics-based end-to-end channel modeling, ray-tracing, and detailed scene geometries with material properties are applied. We consider an urban road with a length of 200 meters and 3 parallel lanes. The lane-changing behavior is not considered, and the total number of vehicles is set to $V=8$.
The tracing data consists of the azimuth and zenith angles of arrival (AoA), azimuth and zenith angles of departure (AoD), time of arrival (ToA), and the delay of multipath. Only the dominant path (i.e., the strongest path) is utilized for each link to characterize the channel propagation features. Tracing data $R[t]\in \mathbb{R}^{V \times V \times 7}$ includes tracing data of V2I and V2V links concatenated in the second dimension. Specifically, the tracing data of V2V links has a dimension of $V \times(V-1) \times 7$, while that of V2I links has a dimension of $V \times 1 \times 7$. We establish a spatial coordinate system with the RSU as the origin at the center of the road, and the location data $L_v[t]\in \mathbb{R}^{V \times 3}$ includes 3D coordinates of each vehicle. The parameter setting of the world model is similar to \cite{hafner2019dream,hafner2023mastering,wang2025dmwm}, and details are presented in Table I. Moreover, all experiments are conducted on a single NVIDIA RTX 3090 GPU, and the training of the world model for wireless networks takes approximately 0.4 GPU days. To better compare the world model-based learning framework with RL approaches, we introduce several state-of-the-art RL baselines including the model-free based discrete soft actor-critic (DSAC) approach \cite{wang2023adaptive} and the model-based policy optimization (MBPO) approach \cite{janner2019trust}.

\begin{figure}[t!]
  \centering
  \includegraphics[scale=0.6]{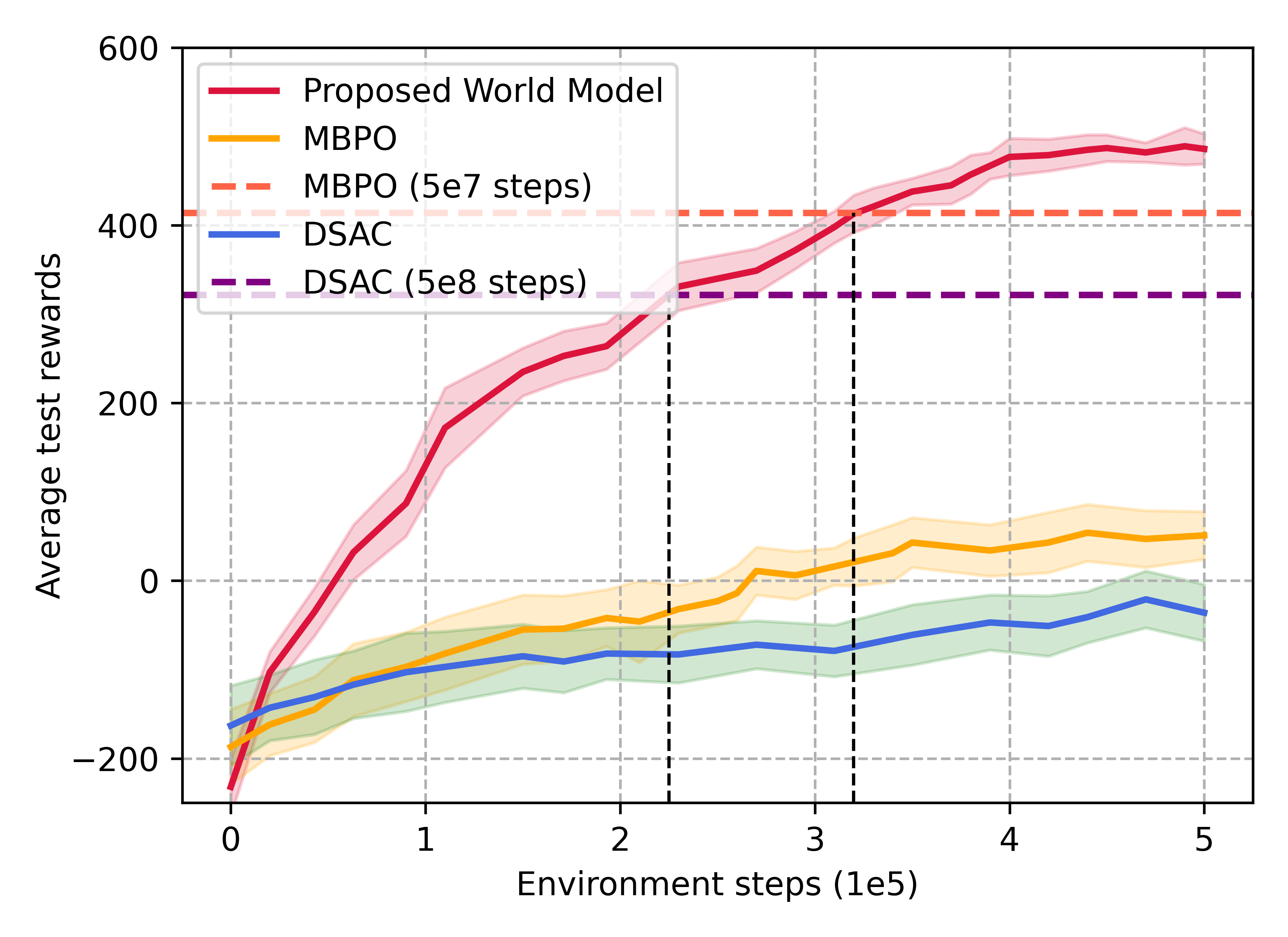}
  \vspace{-0.6cm}
  \caption{The average test rewards under limited environment steps.}
  \vspace{-0.25cm}
\end{figure}

Fig. 4 shows the average test results over 100 test episodes under limited environment steps, where environment steps represent the number of wireless data from the actual V2X network.
The proposed world model-based learning approach exhibits significantly improved data efficiency over the RL baselines, where it achieves superior performance with only $3.5 \times 10^5$ environment steps compared to $5 \times 10^7$ and $5 \times 10^8$ environment steps required by MBPO and DSAC, respectively. Although MBPO exhibits better data efficiency compared to DSAC, it achieves fewer rewards compared to DSAC since it is hard to realize long-term credit assignments. In contrast, DSAC requires an ultra-large amount of environment interactions. The results highlight that the world model overcomes the drawbacks of both MFRL and MBRL. Particularly, the world model accurately captures the dynamics and uncertainty of V2X networks in a compact latent space and trains long-term policies in differentiable imagined trajectories without environment interactions. 

Fig. 5 shows the average CAoI of the V2X network versus different numbers of vehicles. 
The proposed world model-based learning approach achieves 16\% improvement and 26\% improvement, respectively, in terms of CAoI compared to the DSAC approach and the MBPO approach. This is due to the long-term planning ability of the world model that jointly considers CAoI states of vehicles and the reliability of links, thus selecting the optimal solution over a long horizon. It is also observed that the world model with only imagined states, named ``Proposed World Model (Prediction)", can still perform close to the DSAC approach without real-time wireless data. Hence, during intervals when no wireless data can be obtained, the world model can remain capable of making efficient decisions, which is significant for practical deployment and applications. 

\section{Conclusion}
In this paper, we have proposed a novel world model-based learning framework for wireless communication networks. The world model-based approach could overcome the limitations of traditional RL approaches in environment uncertainty, low data efficiency, and long-term planning ability. Taking the highly dynamic mmWave V2X communication network as an example, we have designed a world model that jointly captured the environment dynamics and enabled long-term policy learning through imagined trajectories, without relying on extensive real-time interactions. Moreover, we have utilized the world model's imagination capability to jointly predict the time-varying wireless data and optimize link scheduling in the real-world V2X network. Simulation results show the superiority of the proposed framework and demonstrate significant improvements in data efficiency and the CAoI over state-of-the-art RL baselines. The proposed world model-based learning approach has provided a promising new paradigm for intelligent network management in future wireless networks with complex dynamics and long-term optimization requirements.

\begin{figure}[t!]
  \vspace{-0.5cm}
  \centering
  \includegraphics[scale=0.62]{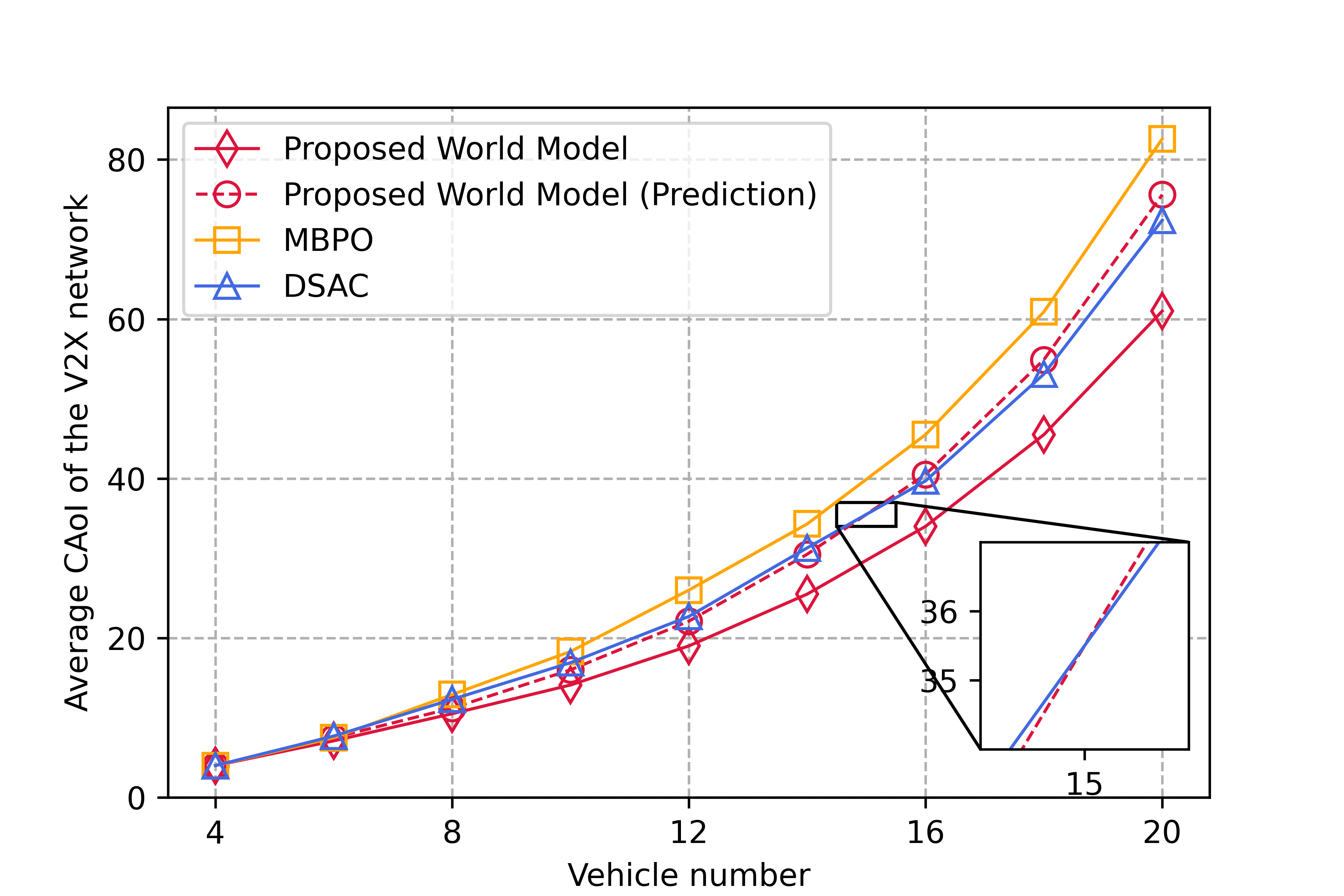}
  \vspace{-0.8cm}
  \caption{The average CAoI of the V2X network versus different number of vehicles.}
  \vspace{-0.3cm}
\end{figure}

\bibliography{IEEEabrv,reference}

\end{document}